# Investigating Market Strength Prediction with CNNs on Candlestick Chart Images

Thanh Nam Duong
FPT Innovation Lab
FPT University
Hanoi, Vietnam
namdthe170744@fpt.edu.vn

Trung Kien Hoang
FPT Innovation Lab
FPT University
Hanoi, Vietnam
kienhthe170770@fpt.edu.vn

Quoc Khanh Duong
FPT Innovation Lab
FPT University
Hanoi, Vietnam
khanhdqhe171671@fpt.edu.vn

Quoc Dat Dinh
FPT Innovation Lab
FPT University
Hanoi, Vietnam
datdqhe171234@fpt.edu.vn

Duc Hoan Le
FPT Innovation Lab
FPT University
Hanoi, Vietnam
hoanldhe170746@fpt.edu.vn

Huy Tuan Nguyen
FPT Innovation Lab
FPT University
Hanoi, Vietnam
tuannhhe170769@fpt.edu.vn

Xuan Bach Nguyen
FPT Innovation Lab
FPT University
Hanoi, Vietnam
bachdxhe176027@fpt.edu.vn

Quy Ban Tran*
Dept of Computing Fundamental
FPT University
Hanoi, Vietnam
bantq3@fe.edu.vn
*Corresponding author

*Abstract*—This paper investigates predicting market strength solely from candlestick chart images to assist investment decisions. The core research problem is developing an effective computer vision-based model using raw candlestick visuals without time-series data. We specifically analyze the impact of incorporating candlestick patterns that were detected by YOLOv8. The study implements two approaches: pure CNN on chart images and a Decomposer architecture detecting patterns. Experiments utilize diverse financial datasets spanning stocks, cryptocurrencies, and forex assets. Key findings demonstrate candlestick patterns do not improve model performance over only image data in our research. The significance is illuminating limitations in candlestick image signals. Performance peaked at approximately 0.7 accuracy, below more complex time-series models. Outcomes reveal challenges in distilling sufficient predictive power from visual shapes alone, motivating the incorporation of other data modalities. This research clarifies how purely image-based models can inform trading while confirming patterns add little value over raw charts. Our content is endeavored to be delineated into distinct sections, each autonomously furnishing a unique contribution while maintaining cohesive linkage. Note that, the examples discussed herein are not limited to the scope, applicability, or knowledge outlined in the paper

## I. Introduction

The financial markets, encompassing a diverse range of assets such as stocks, forex, cryptocurrencies, and commodities, are characterized by their inherent complexity and volatility. Investors and traders seek effective tools and methodologies to analyze market trends and make informed decisions. One such tool, technical analysis, plays a pivotal role in this endeavor by providing insights into the historical and predictive behavior of financial instruments. Among the techniques employed in technical analysis, candlestick patterns have gained prominence for their ability to convey critical information about market sentiment and potential price movements.

Candlestick patterns, a visual representation of price data, offer a powerful means of analyzing market dynamics. These patterns are formed by the open, high, low, and close prices of a given period, allowing traders to interpret and anticipate future market movements. Recognizing the importance of candlestick patterns in technical analysis, this paper addresses the need for advanced computational methods to facilitate their detection and interpretation in a fast-paced financial environment.

The primary purpose of this study is to compare and evaluate the performance of multiple Convolutional Neural Network (CNN) [1,2] deep learning algorithms in the detection of candlestick patterns and, subsequently, in predicting the strength of the trend in various financial markets. To achieve this, we employ a diverse set of deep learning models tailored for different stages of the analysis.

For the detection of candlestick patterns, we explore the effectiveness of state-of-the-art object detection models, including YOLO (You Only Look Once) [3] and Faster R-CNN [4] (Region-based Convolutional Neural Networks). Additionally, we experiment with techniques that convert candlestick chart images into time series data, allowing us to leverage specialized libraries and tools to further enhance pattern recognition

In terms of predicting the strength of market trends, we employ a combination of deep learning models, including traditional CNNs with different architectures and a novel approach based on the Deep Candlestick Predictor (DCP) [5] framework, which includes the Decomposer method, CNN-Autoencoder (CAE) analysis, and 1D Convolutional Neural Networks (CNN1D) [5]. This multifaceted approach provides a comprehensive understanding of the market trend's vigor, enhancing our ability to make informed investment decisions, as inspired by the Deep Candlestick Predictor framework for forecasting price movements from candlestick charts. Additionally, we propose an innovative approach for leveraging candlestick patterns as supplementary information. This involves the extraction of relevant features using a distinct CNN architecture, which is then fused with the main CNN through dense layers. This novel technique enhances the model's ability to capture and utilize candlestick pattern information for more accurate predictions and trend analysis.

Crucially, our analysis encompasses different financial markets and assets, spanning various goods in different countries. This broad dataset includes financial instruments such as Bitcoin, forex currency pairs, and stocks, offering a well-rounded examination of the applicability and generalizability of our deep learning models across diverse market scenarios. By comparing these models across different datasets, we aim to provide insights into their versatility and robustness in predicting market trends.

In summary, this paper contributes to the ongoing efforts to leverage artificial intelligence and deep learning techniques for more accurate and efficient financial market predictions. By comparing the performance of multiple deep learning models on diverse datasets, we seek to advance the field of technical analysis and enhance traders' and investors' ability to make data-driven decisions in the dynamic world of financial markets.

## II. RELATED WORK

The use of candlestick patterns in financial analysis and prediction has garnered significant attention in the economic research community. This section provides an overview of relevant literature and research efforts related to candlestick pattern detection and market trend prediction.

### A. Candlestick Pattern Detection

The study of candlestick patterns, deeply rooted in technical analysis, has traditionally relied on manual inspection by traders and analysts. Recent advancements in computer vision and deep learning have introduced automated methods for candlestick pattern detection.

Jun-Hao Chen et al.'s work introduced a dynamic deep convolutional candlestick learner, which leverages a modified YOLO architecture for localizing and recognizing patterns from candlestick charts[6]. Their approach is focused on using object detection techniques to detect and classify candlestick patterns efficiently.

On the other hand, Kusuma et al.'s work [7] uses simple Convolutional Neural Networks (CNNs) to detect candlestick patterns. Their method is tailored for pattern recognition and market trend prediction. Kietikul Jearanaitanakij and Bundit Passaya, in their work [8] also adopted a CNN architecture to extract features from candlestick charts for pattern detection. Their results show favorable performance compared to alternative detection architectures like ResNet18.

While others use the standalone method, others such as Marc Velay et al's work [9] combine methods to get better results such as using CNN to extract features from the candlestick and LSTM to recognize the pattern within.

### B. Market Trend Prediction

The prediction of market trends and price movements is a fundamental goal in finance analysis. Deep learning models, with their ability to capture complex patterns in financial data, have emerged as powerful tools for this purpose.

Some of the noticeable works from Wang et al [10], Mu et al [11], Can Yang et al [12], Kim T et al [13], Yang Liu [14], and Sidra Mehtab et al [15] are all explore the possibility of combining convolutional network such as CNN with some NLP modules like the LSTM architecture to predict the price movement in a period.

On the contrary, Siou Jhih Guo et al [16] and Chih-Chieh Hung et al [17] proposed a Deep candlestick Predictor (DCP) framework to forecast stock movement which has been shown to have high proficiency. Chih-Chieh Hung et al [18] also presented the Deep predictor for price movement (DPP) framework capable of forecasting the change in the stock market using CNN and RNN combination.

### C. Cross-Market Anaalysis

Our research builds upon the existing literature by extending the analysis to a diverse set of financial markets and assets and incorporating advanced object detection techniques. We examine various financial instruments, including Bitcoin, forex currency pairs, and stocks, to assess the adaptability and generalizability of deep learning models across different market scenarios.

This cross-market analysis aligns with the findings of Abdellah EL ZAAR et al [14], who explored Ethereum cryptocurrency entry points and trend prediction using Bitcoin correlation and multiple data combinations.

In summary, the literature review highlights the evolution of candlestick pattern detection and market trend prediction techniques, with a specific focus on the use of Convolutional Neural Networks (CNNs). Our research contributes to this body of LSTM work by comparing and evaluating the performance of multiple deep learning models on diverse datasets, aiming to advance the field of technical analysis and empower traders and investors with data-driven decision-making tools in the dynamic realm of financial markets.

## III. METHOD

In this section, we will furnish a comprehensive account of the data employed in our study, elucidating the intricacies of data collection and processing. Additionally, we will expound upon the models adopted in our research, providing detailed insights into their implementation.

### A. Data

*1) Data collection*

To gather necessary historical data, we accessed the financial market's API provided by Yfinance Exchange. We selected stocks(AAPL), exchange(EUR_USD), and crypto(Bitcoin) as our trading instruments of interest to get data for our train model and perform predictions. The time data we collected is from 2017 up until 30 October 2023. The reason behind choosing these is that it change frequently, making it an excellent data for training. Utilizing the API, we achieved daily price with five elements, these are: Date, along with Open, Close, Min, and Max price.

The decision to opt for a daily time frame was influenced by the primary object of detecting medium-term trends while keeping the computational effort under control. This period achieves an optimal mix between catching intraday variations and offering enough data points for significant pattern detection. Indeed, using a daily timeline is appropriate, especially for such a cryptocurrency like Bitcoin, which was created on January 1, 2017, and has been around for approximately six years today.

*2) Data processing*

Subsequently, a specialized custom function is employed to identify specific candlestick patterns that effectively define the strength of impending trends within the cryptocurrency market. These patterns are then labeled as either strong or weak, providing a vital ground truth for the model's learning process.

We use Talib library[19] to detect the existence of patterns in our data. Then whenever there is a pattern, select those pattern candlesticks with the past (30-number of candle) candles to feed into our function to detect whether the upcoming trend is strong or weak. Then to facilitate the model's understanding of these labeled candlestick patterns, a Python library is employed to transform this data into image representations, this will include the image for the pattern and the image for its history. Additionally, the pattern image needs to be annotated with our custom image processing. This conversion allows for a visual interpretation of these patterns, making it more comprehensible for the Convolutional Neural Network (CNN). The images, paired with their corresponding labels, are then fed into the CNN model, which is designed to learn and extract intricate features from these visual representations

So the output will have the label of the pattern appear along with its history image, pattern image, pattern label, and label for the upcoming strength

*B. Method*

*1) DCP*

Our approach builds upon the framework introduced in the paper "Deep Candlestick Predictor," with tailored adaptations and modifications to address the specific requirements of our work. In this section, we present the key components of our methodology, including the Decomposer, the CNN-autoencoder, and the CNN1D, emphasizing the distinctive changes introduced.

*a) Decomposer*

Our inspiration for the Decomposer concept is drawn from the referenced "Deep Candlestick Predictor" paper. However, detailed specifics on the decomposition process were not provided in the original paper. To adapt this concept for our work, we utilized computer vision techniques to implement a custom Decomposer. This module is designed to segment and extract sub-charts from the original candlestick chart images. By doing so, we focus on individual sub-charts, facilitating subsequent models in identifying and analyzing candlestick patterns.

*b) CNN-Autoencoder*

In the realm of candlestick pattern analysis, a significant alteration has been made to the CNN-autoencoder. In the original paper, the autoencoder was designed for single-channel (black and white) images. However, we modified the architecture to accommodate three-channel RGB images. This adaptation preserves and utilizes color information when analyzing sub-charts extracted by the Decomposer, enhancing the richness of data available to our model and potentially improving its ability to detect and interpret candlestick patterns.

*c) CNN1D*

Adjustments were made to the CNN1D model to optimize its performance. In the original architecture, a specific number of layers were defined. To address a potential issue leading to a zero-size output, we reduced the number of layers by half. This modification aims to maintain the model's ability to process and analyze data effectively while avoiding architectural constraints.

By building upon the foundational concepts from the "Deep Candlestick Predictor" paper and customizing them to our specific needs, our methodology incorporates innovative techniques for Decomposition, integrates color information in the CNN-autoencoder, and ensures the practicality and efficiency of the CNN1D model. These modifications are designed to enhance our overall approach and improve the accuracy and robustness of candlestick pattern detection and market trend analysis.

*2) CNN*

*a) Two-stream CNN*

Furthermore, in our research, we strategically incorporated an alternative methodology known as the Two-Stream Convolutional Neural Network (CNN). This distinctive approach is characterized by the integration of two distinct streams, each uniquely designed to fulfill specific purposes within the framework of our study.

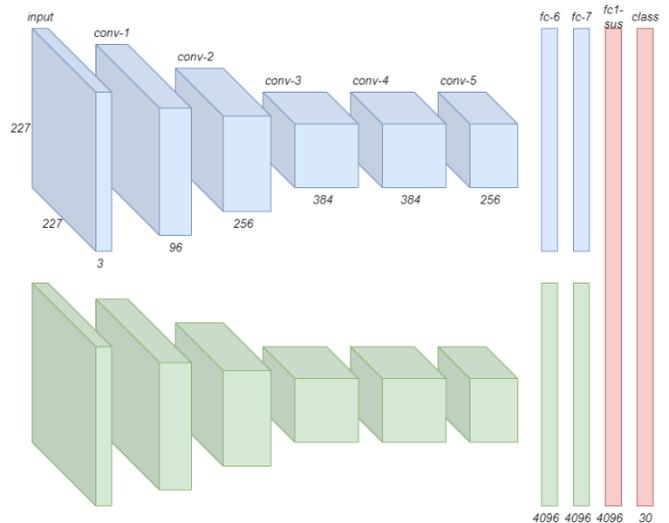

Fig. 1. Two stream CNN architecture

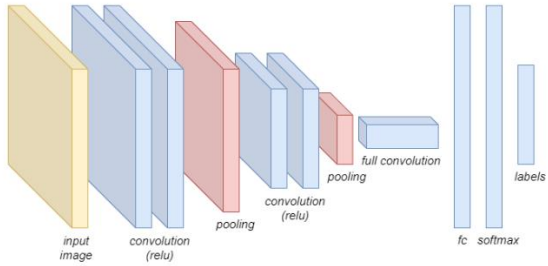

Fig. 2. Simple CNN architecture

In the main stream, we employ a CNN architecture to extract features directly from the candlestick chart images. This stream focuses on capturing information from the candlestick charts, allowing us to analyze the primary data source comprehensively. The features extracted from this stream provide essential insights into the underlying market trends

The pattern stream is dedicated to extracting features specifically from the candlestick patterns present in the chart. This stream uses its own CNN architecture, tailored to identify and interpret candlestick patterns effectively. By isolating the patterns from the main chart, we can focus on this critical element of technical analysis.

The features extracted from both streams are then combined in a fully connected layer. This merging of information from the main chart and the candlestick patterns allows us to leverage the strengths of both aspects, enhancing the overall accuracy and predictive power of our model. This innovative Two-Stream CNN approach provides a holistic view of the market dynamics, taking into account both the broader market context and the specific candlestick patterns present, contributing to more informed investment decisions.

*b) Simple CNN*

The model architecture used was a convolutional neural network (CNN) with a variety of backbone architectures, including ResNet, VGG, and InceptionNet. The CNN consisted of a series of convolutional layers, followed by a pooling layer and a fully connected layer. The output of the fully connected layer was a single value, which represented the predicted strength of the market.

## IV. RESULTS

In this section, we present the results of our research on evaluating the strength of candlestick charts for predicting market conditions. We conducted two major experiments: Pure CNN and Decomposer. The Pure CNN approach involved both "Non-Pattern" and "Include-Pattern" variants, where we either fed the model raw candlestick images or images with detected patterns. The Decomposer experiments only on raw candlestick images. Across all experiments, we evaluated performance using three key metrics: accuracy, F1 score, and AUC. These metrics quantify model capabilities on the candlestick prediction task. The experiments aim to ascertain whether candlestick patterns provide useful signals for forecasting market strength. By comparing Pure CNN to Decomposer, we can determine the value of explicitly extracting pattern information versus using only raw chart images.

TABLE I. PERFORMANCE OF DIFFERENT MODELS ON APPL TEST DATASET

| Model | Accuracy | F1-Score | AUC |
|---|---|---|---|
| EfficientNet_B0 | 0.638 | 0.63 | 0.632 |
| ResNet18 | 0.623 | 0.623 | 0.626 |
| VGG11 | 0.646 | 0.645 | 0.646 |
| **VGG16*** | **0.701** | **0.699** | **0.699** |

a. *mean the best

To determine the optimal model for extracting meaningful features from candlestick charts, we surveyed several state-of-the-art deep learning architectures. Based on experiments using candlestick data, we found that VGG16 emerged as the best performer for feature extraction. Compared to other models like ResNet50, VGG16 more effectively learned representations capturing informative patterns in the candlestick visualizations. The key advantages of VGG16 are its use of small convolutional filters to identify localized features and its depth to learn hierarchical abstractions. Given these findings, we utilized VGG16 as the foundation for all subsequent experiments to extract expressive feature representations from the candlestick charts before predicting market strength. The feature maps generated by VGG16 contain activations highlighting the most salient visual patterns and shapes within the candlestick images. By leveraging VGG16, our model is equipped to interpret the nuances of candlestick charts when evaluating the state of the market.

In our first major experiment, we employed a Pure CNN with an "include-pattern" approach where candlestick patterns were directly incorporated into the model. Specifically, we utilized YOLOv8 to detect patterns in the candlestick charts and VGG16 as the backbone to extract visual features. A key training technique we leveraged was "Force Teaching," where we fed the model the exact pattern images rather than relying solely on YOLOv8's detected outputs. This accounts for imperfect accuracy in pattern detection that could otherwise negatively impact learning for strength classification. We evaluated the "include-pattern" method against our "non-pattern" baseline on various datasets - the EUR_USD exchange rate, BTC_USD crypto price, and AAPL stock. Across all datasets, we found the "include-pattern" approach did not outperform the "non-pattern" and in some cases was significantly worse. The direct usage of candlestick patterns failed to boost predictive performance over pure price data alone. Our results suggest the "include-pattern" methodology is not an effective avenue for this task. Therefore, we do not recommend the "include-pattern" approach for predicting market strength from candlestick charts.

TABLE II. WITH PATTERN PERFORMANCE OF VGG16 ON DATASETS

| Dataset | Test accuracy | Test F1 | Test AUC |
|---|---|---|---|
| BTC-USD | 0.792 | 0.787 | 0.793 |
| EUR-USD | 0.623 | 0.594 | 0.593 |
| APPL | 0.653 | 0.642 | 0.645 |

TABLE III. WITHOUT PATTERN PERFORMANCE OF VGG16 ON DATASETS

| Dataset | Test accuracy | Test F1 | Test AUC |
|---|---|---|---|
| BTC-USD | 0.765 | 0.757 | 0.758 |
| EUR-USD | 0.753 | 0.733 | 0.729 |
| APPL | 0.701 | 0.699 | 0.699 |

We also implement the Pure CNN model on "non-pattern" - raw candlestick chart images. Using data from stocks, cryptocurrencies, and forex, we evaluated performance both on individual assets and on aggregated datasets combining multiple assets within each market type. Across all trials, predictive accuracy, F1 scores, and AUC consistently reached approximately 0.7 regardless of whether we used data from a single asset or a diverse blend of assets. The consistent metrics indicate the pure CNN achieved stable performance for predicting market strength solely from the candlestick visuals, without relying on detected patterns. However, we did not observe any significant improvement from expanding the data diversity.

Our next main experiment is using Decomposer with raw candlestick images. Within the Decomposer approach, we deploy this model exclusively within the "Non-Pattern" framework. This choice is made because the candlestick charts in the DCP approach have already been segmented into sub charts, each consisting of three candlesticks, which partially reveal the inherent pattern. To elucidate the training and performance aspects, we include a table depicting the training progress of the DCP model and a separate table presenting the classification results achieved through the DCP approach.

After subjecting the DCP model to extensive training and evaluation, it becomes evident that certain metrics, particularly the F1 score and accuracy, remain persistently stagnant at approximately 0.5 for a considerable span of 20 epochs. This performance is notably suboptimal, as it closely resembles the outcome of a random guess with a 50% probability for each event. In essence, this suggests that the DCP model, under the current configuration, is no more informative or accurate than a random selection.

TABLE IV. VGG16 PERFORMANCE ON SINGLE DATASET

| Dataset | Accuracy | F1-Score | AUC |
|---|---|---|---|
| BTC | 0.765 | 0.757 | 0.758 |
| ETH | 0.655 | 0.616 | 0.635 |
| EUR-USD | 0.753 | 0.733 | 0.729 |
| USD-CAD | 0.763 | 0.723 | 0.749 |
| AMZN | 0.723 | 0.719 | 0.719 |
| APPL | 0.701 | 0.699 | 0.699 |
| AMD | 0.732 | 0.73 | 0.738 |

TABLE V. VGG16 PERFORMANCE ON MERGE DATASET

| Dataset | Accuracy | F1-Score | AUC |
|---|---|---|---|
| BTC ETH | 0.771 | 0.76 | 0.775 |
| EUR-USD USD-CAD | 0.71 | 0.688 | 0.689 |
| AMZN APPL AMD | 0.697 | 0.697 | 0.701 |

## V. DISCUSSTION

In the ensuing discussion, we delve into the significance of our research findings, drawing comparisons with antecedent studies. We address the limitations inherent in our study and proffer recommendations for future research endeavors.

### A. Pattern vs Non-Pattern

In the comparative analysis of the "Include-Pattern" and "Non-Pattern" experiments, it becomes evident that the former, which employed YOLOv8 for pattern detection and VGG16 for feature extraction, did not consistently outperform the "Non-Pattern" approach and, in some cases, even exhibited poorer performance. The use of strategy, where exact pattern images were employed for training, may have contributed to this inconsistency. It is crucial to delve into the reasons behind this discrepancy in performance, as it may be attributed to potential overfitting issues or the inherent limitations of the YOLOv8 and VGG16 combination. Consequently, the "Include-Pattern" approach is not recommended for predicting market strength from candlestick charts, raising questions about the viability of such a strategy in financial market analysis.

### B. Image vs Time-serires

In the comparison between time-series and image data for candlestick pattern detection using CNNs, our findings unequivocally endorse the superiority of time-series data. Candlestick charts inherently embody temporal dynamics crucial for pattern identification. Time-series data preserves this sequential information, allowing CNNs to effectively capture evolving market patterns. In contrast, converting candlestick patterns into images sacrifices temporal context, hindering the model's ability to discern nuanced relationships between consecutive candles. The adaptability of time-series data across diverse time intervals enhances its utility in various trading strategies, contributing to better generalization and robustness. Moreover, detecting patterns in time-series results better because of the rule-based if-else detection while CNN doesn't have.

### C. Practical Implications

The practical implications of integrating using CNN for candlestick pattern prediction in financial markets are multifaceted and hold significant promise for traders and investors. By automating the identification and interpretation of candlestick patterns, CNNs have the potential to significantly enhance technical analysis, streamlining the decision-making process. This data-driven approach offers traders and investors the ability to make more informed decisions, leveraging historical price data and market behavior for greater precision.

Moreover, CNNs operate with efficiency and speed, a crucial advantage in rapidly changing markets, where timely decisions can make a substantial difference. Their capacity for risk mitigation is also noteworthy, as accurate predictions serve as early warnings of potential trend reversals or market movements, empowering traders to execute effective risk management strategies. The integration of CNN-based predictions into algorithmic trading strategies provides an avenue for automated, data-backed trading decisions, potentially capitalizing on short-term price movements. Furthermore, investors can utilize these predictions to diversify their portfolios effectively across different markets, reducing risk and optimizing returns. The adaptability of CNN models through continuous learning ensures that they can stay ahead of evolving market conditions and patterns. This integration of CNNs into trading platforms makes data-driven analysis readily accessible, facilitating the incorporation of advanced insights into traders' strategies. These practical implications collectively offer market participants a powerful tool for more informed and efficient decision-making, potentially leading to improved trading outcomes and enhanced risk management in the dynamic landscape of financial markets.

### D. Limitations

Primarily, Convolutional Neural Networks (CNNs) are contingent upon the quality and quantity of the training dataset. The efficacy of these networks can be compromised when historical data is limited or tainted by noise, thereby impeding their ability to generalize optimally to authentic market conditions. Additionally, the ubiquitous machine learning challenge of overfitting exacerbates the situation. In instances where the model architecture exhibits excessive complexity or where the available data is insufficient, CNNs may encounter difficulties in delivering robust performance when confronted with unseen market data, thereby engendering predictions of questionable reliability.

### E. Future works

In the realm of future research endeavors, an imperative focus lies on the cultivation of more expansive and intricate datasets. The foundational importance of high-quality, diversified datasets cannot be overstated, as they constitute a cornerstone for the effective training of Convolutional Neural Network (CNN) models. Subsequent investigations should be directed towards the development of datasets that encapsulate a broader spectrum of market conditions, incorporating diverse asset classes, temporal intervals, and levels of market volatility. This enriched dataset should not only extend across varied financial markets but also integrate a multitude of historical data points, thereby enhancing its representational fidelity with respect to real-world trading scenarios.

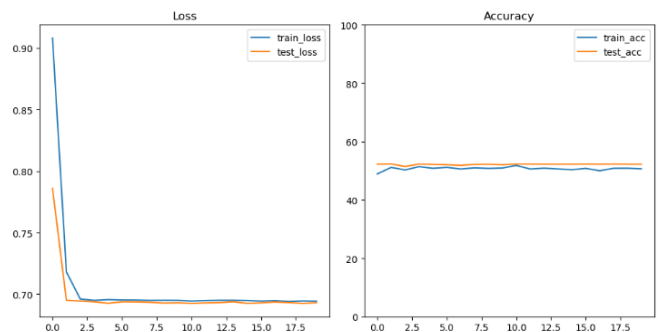

Fig. 3. Loss and accuracy of DCP model

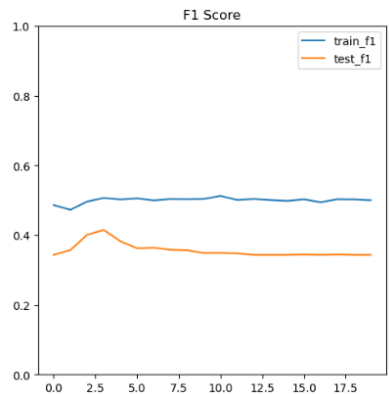

Fig. 4. F1-Score of DCP model

Furthermore, there exists a necessity to engage in systematic experimentation with and refinement of diverse methodological approaches, aimed at attaining heightened accuracy scores. Such approaches encompass potential variations in model architectures, training strategies, and data preprocessing techniques. One promising avenue for exploration is the application of transfer learning, wherein pre-trained models undergo fine-tuning specifically tailored for candlestick pattern prediction. This avenue holds promise for optimizing model performance.

Moreover, the incorporation of advanced techniques merits investigation for their potential to augment model efficacy. This includes the exploration of attention mechanisms and the integration of recurrent neural networks (RNNs) in conjunction with CNNs. Such synergistic approaches have the potential to significantly enhance model performance, particularly in capturing prolonged trends and intricate market dynamics. The pursuit of these avenues represents a promising trajectory for advancing the capabilities of CNN-based models in the realm of financial market prediction.

### VI. CONCLUSION

In conclusion, this paper has presented a comprehensive comparative analysis of multiple CNN deep learning architectures for the detection of candlestick patterns and the subsequent prediction of market trend strength. The models used in this study encompassed YOLO, Faster R-CNN, a method that converts candlestick chart images into time series data for pattern detection, and CNNs along with the Deep Candlestick Predictor (DCP) framework for strength prediction.

Our findings have revealed that YOLO exhibited the highest accuracy, reaching 80% in detecting candlestick patterns. However, for the prediction of market trend strength, a standard CNN architecture without the use of additional pattern information proved to be sufficient, indicating the importance of pattern detection in the trading process but also the potential for more streamlined models in trend analysis.

As a basis for future research, we suggest further enhancing the predictive capabilities of these models by incorporating Long Short-Term Memory (LSTM) networks to extract more relevant information from the data. Additionally, expanding the scope of input data beyond candlestick patterns to encompass factors such as trading volume, moving averages, Ichimoku Kinko Hyo, and other technical indicators may provide a more comprehensive and accurate basis for financial market predictions. These proposed enhancements are expected to contribute to a deeper understanding of market dynamics and the development of more robust predictive models in the field of financial market analysis.

**IEEE conference templates contain guidance text for composing and formatting conference papers. Please ensure that all template text is removed from your conference paper prior to submission to the conference. Failure to remove template text from your paper may result in your paper not being published.**

Authors' background (This form is only for submitted manuscript for review) *This form helps us to understand your paper better, the form itself will not be published. **Please delete this form on final papers.** *Title can be chosen from: master student, Phd candidate, assistant professor, lecture, senior lecture, associate professor, full professor.

| Your Name | Title* | Affiliation | Research Field | Personal website |
|---|---|---|---|---|
|  |  |  |  |  |
|  |  |  |  |  |